\pdfoutput=1

\documentclass[11pt]{article}

\usepackage{hyperref}
\usepackage[]{EMNLP2022}
\usepackage{times}
\usepackage{placeins}
\usepackage{latexsym}
\usepackage{comment}
\usepackage{longtable}
\usepackage{graphicx}
\usepackage{amsmath}
\usepackage{multirow}
\usepackage{adjustbox}
\usepackage{tabularx}

\usepackage[T1]{fontenc}

\usepackage[utf8]{inputenc}

\usepackage{microtype}

\usepackage{inconsolata}

\title{(Psycho-)Linguistic Features Meet Transformer Models for Improved Explainable and Controllable Text Simplification}

\author{
    Yu Qiao$^1$, Xiaofei Li$^1$, Daniel Wiechmann$^2$, Elma Kerz$^1$ \\
    $^1$ RWTH Aachen University \\
    $^2$ University of Amsterdam \\
    \texttt{\{yu.qiao, xiaofei.li1\}@rwth-aachen.de}\\
    \texttt{d.wiechmann@uva.nl, elma.kerz@ifaar.rwth-aachen.de}
}

\begin{document}
\maketitle
\begin{abstract}
State-of-the-art text simplification (TS) systems adopt end-to-end neural network models to directly generate the simplified version of the input text, and usually function as a blackbox. Moreover, TS is usually treated as an all-purpose generic task under the assumption of homogeneity, where the same simplification is suitable for all. In recent years, however, there has been increasing recognition of the need to adapt the simplification techniques to the specific needs of different target groups. In this work, we aim to advance current research on explainable and controllable TS in two ways: First, building on recently proposed work to increase the transparency of TS systems \cite{garbacea-etal-2021-explainable}, we use a large set of (psycho-)linguistic features in combination with pre-trained language models to improve explainable complexity prediction. Second, based on the results of this preliminary task, we extend a state-of-the-art Seq2Seq TS model, ACCESS \cite{martin-etal-2020-controllable}, to enable explicit control of ten attributes. The results of experiments show (1) that our approach improves the performance of state-of-the-art models for predicting explainable complexity and (2) that explicitly conditioning the Seq2Seq model on ten attributes leads to a significant improvement in performance in both within-domain and out-of-domain settings.
\end{abstract}

\section{Introduction}

Text simplification (henceforth TS) is a natural language generation task aimed at transforming a text into an equivalent that is more readable and understandable for a target audience, while preserving the original information and underlying meaning. It  involves a number of transformations applied at different linguistic levels, including lexical, syntactic and discourse aimed at reducing the complexity of content for the purpose of accessibility and readability  \cite[see][for overviews]{siddharthan2011text,shardlow2014survey,alva2020data,al2021automated,jin2022deep}. Simplification techniques have been shown to be beneficial as reading supports across a wide range of populations, from children \cite{de2010text, kajiwara2013selecting}, individuals with language disorders such as aphasia \cite{carroll1999simplifying, devlin2006helping}, dyslexia \cite{rello2013simplify,rello2013impact} or autism \cite{evans2014evaluation}; language learners and non-native English speakers \cite{petersen2007text,paetzold2016unsupervised}, and people with low literacy skills \cite{max2006writing, candido2009supporting,watanabe2009facilita}. 
Moreover, TS techniques have also been successfully employed as a preprocessing step to improve the performance of various downstream NLP tasks such as parsing \cite{chandrasekar1996motivations}, machine translation \cite{gerber1998improving,hasler2017source}, summarization \cite{beigman2004text,silveira2012combining}, semantic role labeling \cite{vickrey2008sentence}, and information extraction \cite{miwa2010entity}.
TS approaches typically learn simplification transformations using parallel corpora of matched original and simplified sentences and can be classified into six categories \cite[for recent overviews see][]{alva2020data,al2021automated}: Early approaches relied on either (1) manually generated rules for splitting and reordering sentences \cite{candido2009supporting,siddharthan2011text} or (2) learned simple lexical simplifications, i.e., one-word substitutions \cite{devlin1998use,carroll1998practical}. Subsequent work has introduced (3) phrase-based and syntax-based statistical machine translation techniques \cite{wubben2012sentence,xu2016optimizing}, (4) grammar induction \cite{paetzold2013text,feblowitz2013sentence}, and (5) semantics-assistance, i.e., obtaining semantic representations of the original sentences \cite{narayan2014hybrid, vstajner2017leveraging}.
More recently, TS tasks have been approached with (6) neural machine translation methods, in particular sequence-to-sequence (Seq2Seq) models using an attention-based encoder-decoder architecture \cite{nisioi2017exploring,alva2017learning,Zhang2017ACS}.
While the performance of Seq2Seq TS models is impressive, most of these models are black-box models characterized by the lack of interpretability of their procedures \cite{alva2020data}. In recent years, there have been growing calls to a move away from black-box models toward explainable (white-box) models \cite{loyola2019black, qiao2020language,aguilar2022towards}. Moreover, recent work in TS suggests that the performance of state-of-the-art  TS systems can be improved by conducting explainable complexity prediction as a preliminary step \cite{garbacea-etal-2021-explainable}. 

\vspace{-1.5mm}
Another important trend in current TS research is the growing recognition that the concept of ‘text complexity’ is not homogeneous for different target populations \cite{gooding-etal-2021-word}. That is, rather than viewing TS as a general task where the same simplification is appropriate for everyone (one-fits-all approach), researchers are placing a greater emphasis on the need to develop TS systems that can flexibly adapt to the needs of different audiences: For example, while second language learners might struggle with texts with rare or register-specific vocabulary, aphasic patients might be overwhelmed by a high cognitive load associated with long, syntactically complex sentence structures. In response, recent TS research has begun to adopt methods proposed in controllable text generation research (see the \ref{related} section for further discussion). Controllable text generation refers to the task of generating text according to a given controlled property of a text. More generally, the development of controllable text generation systems makes an important contribution to the general development of ethical AI applications. This requires the ability to avoid biased content such as gender bias, racial discrimination, and toxic words. In addition, it is widely seen as critical to the development of advanced text generation technologies that better address specific needs in real-world applications \cite{prabhumoye2020exploring,zhang2022survey}. For example, the task of dialog response generation requires effective control over text attributes associated with emotions \cite{li2021emoelicitor} and persona \cite{zhang2018personalizing}. In the context of TS, the relevant attributes involve various linguistic aspects of text complexity \cite{siddharthan2011text}. By combining multiple attributes, a natural language generation system can theoretically achieve not only greater controllability but also greater interpretability. This requires the inclusion not only of surface features, but also of more sophisticated features.  Traditionally, TS has used readability measures that consider only surface features. For example, the Flesch Reading Ease Score \cite{flesch1948new}, a commonly used surface feature, measures the length of words (in syllables) and sentences (in words). While readability has been shown to correlate to some degree with such features \cite{just1980theory}, there is general consensus that they are insufficient to capture the full complexity of a text. 

In a nutshell, despite significant progress in data-driven text simplification, the development of explainable and controllable models for automatic text simplification remains a challenge. In this paper, we advance current research on explainable and controllable text simplification in two ways:

\vspace{-2mm}
\begin{enumerate}
    \item First, we use what is, to our knowledge, the most comprehensive set of (psycho-)linguistic features that goes beyond traditional surface measures and includes features introduced in the recent literature on human (native and non-native) language learning and processing.  These encompass lexical, syntactic, register-specific ngram, readability and psycholinguistic features and are used in combination with pre-trained language models to improve explainable complexity prediction proposed in \citet{garbacea-etal-2021-explainable}.
    \vspace{-2mm}
    \item Second, based on the results of this preliminary task, we extend a state-of-the-art Seq2Seq TS model, ACCESS \cite{martin-etal-2020-controllable}, to provide explicit control over ten attributes so that simplifications can be adapted to the linguistic needs of different audiences.
\end{enumerate}
\vspace{-2mm}

The remainder of the paper is organized as follows: Section 2 provides a concise overview of related work in the field of explainable and controllable text generation with a focus on TS. Section 3 outlines the experimental setup including the description of three benchmark datasets used (Section 3.1), the type of features extracted from these datasets (Section 3.2), and the models performed to improve explainable and controllable TS (Sections 3.3-3.5). Section 4 presents and discusses the results of our experiments before presenting conclusions and future work in Section 5. Sections 6 and 7 address the limitations of the study and point out ethical considerations.

\vspace{-2mm}
\section{Related work} \label{related}
\vspace{-2mm}
State-of-the-art systems for controllable text generation typically use a Sequence-to-Sequence (Seq2Seq) architecture. These systems follow either a learning-based or a decoding-based approach: In the learning-based approaches, the Seq2Seq model is conditioned on the attribute under consideration at training time and then used to control the output at inference time. Within this approach, controlled text generation can be achieved by disentangling the latent space representations of a variational autoencoder between the text representation and the controlled attributes \cite{hu2017toward}. Decoding-based methods, on the other hand, are based on a Seq2Seq training setup that is modified to control specific attributes of the output text \cite{kikuchi2016controlling, scarton2018learning}. For instance, \citet{kikuchi2016controlling} controlled the length of the text output in the encoder-decoder framework by preventing the decoder from generating the end-of-sentence token before the desired length was reached, or by selecting only hypotheses of a certain length during the beam search. Recently, \citet{martin-etal-2020-controllable} adapted a discrete parameterization mechanism to the task of sentence simplification by conditioning on relevant attributes. Building on the earlier work of TS \cite{scarton2018learning}, their model, called ACCESS -- short for AudienCe-CEntric Sentence Simplification -- provides explicit control of TS by conditioning the output returned by the model on specific attributes. These attributes and their values are prepended as additional inputs to the source sentences at train time as plain text `parameter tokens'. Results of experiments on the WikiLarge corpus \cite{zhang2017sentence} show that with carefully chosen values of three attributes - (i) character length ratio between source sentence and target sentence, (ii) normalized character-level Levenshtein similarity between source and target, and (iii) WordRank, a proxy to lexical complexity, the ACCESS model outperformed previous TS systems on simplification benchmarks, achieving state-of-the-art at 41.87 SARI, corresponding to a +1.42 improvement over the best previously reported score. 

Another recently introduced line of research, on which the present work builds, explores how the transparency and explainability of the TS process can be facilitated by decomposing the task into several carefully designed subtasks. More specifically, \citet{garbacea-etal-2021-explainable} propose that TS benefits from a preparatory task aimed at the explainable prediction of text complexity, which in turn is divided into two subtasks: (1) classifying whether a given text needs to be simplified or not (complexity prediction) and (2) highlighting the part of the text that needs to be simplified (complexity explanation). \citet{garbacea-etal-2021-explainable} focuses on empirical analysis of the two subtasks of explicable prediction of text complexity. Specifically, they conduct experiments using a broad portfolio of deep and shallow classification models in combination with model-agnostic explanatory techniques, in particular LIME \cite{ribeiro2016model} and SHAP \cite{lundberg2017unified}. The results of their experiment show that a combination of a Long Short-Term Memory network at the word level and LIME explanations can achieve strong performance on datasets. As a next step, they conduct follow-up experiments with state-of-the-art controllable end-to-end text generation systems, including ACCESS. The results of these experiments suggest that the performance of state-of-the-art TS models can be significantly improved in out-of-sample text simplification simply by applying explainable complexity prediction as a preliminary step.

\vspace{-2mm}
\section{Experimental Setup}
\vspace{-2mm}

In this section, we first introduce the three datasets used in our experiments (Section \ref{datasets}) and the type of (psycho-)linguistic features used in our models (Section \ref{cocogen}). We then describe the methods used to address the three subtasks, i.e., (1) complexity prediction, (2) complexity explanation, and (3) simplification generation. For subtask (1), we perform experiments with five complexity prediction models described in Section \ref{complexity prediction}: (1) A word-level Long Short-Term Memory (LSTM) network, (2) a fine-tuned pre-trained BERT-based model, (3) and (4) two hybrid Bidirectional Long-Term Memory (BLSTM) classifiers that integrate GloVe word embeddings with (psycho-)linguistic features using different fusion methods, and (5) A hybrid classifier that integrates the those features with BERT representations. In subtask (2), we apply these five models to identify the complex parts of a given input set to facilitate model validation and evaluation (section \ref{complexity explanation}). In Section \ref{simplification generation}, we turn to subtask (3) and introduce an extended ACCESS model, which we refer to as ACCESS-XL, containing a total of ten control features (parameter tokens) covering several dimensions of linguistic complexity.

\subsection{Datasets} \label{datasets}

We conducted our experiments with three benchmark datasets and ground truth complexity labels that were also used in \citet{garbacea-etal-2021-explainable}: (1) the WikiLarge corpus \citet{Zhang2017ACS}, composed of parallel-aligned "Wikipedia-simple-Wikipedia" sentence pairs, (2) the Newsela corpus \cite{xu2015problems}, comprised of news articles simplified by professional news editors, and (3) the Biendata dataset, comprising matches of research papers from different scientific disciplines with press releases describing them\footnote{https://www.biendata.com/competition/
hackathon}. The size of the three datasets and their distribution among training, validation, testing datasets are shown in Table 1.

\begin{table}[h!]
\begin{tabular}{|l|l|l|l|}
\hline
\textbf{Dataset}   & \textbf{Training} & \textbf{Validation} & \textbf{Test}  \\ \hline
Newsela   & 94,944     & 1,131        & 1,079  \\ \hline
WiKiLarge & 207,480    & 30,632       & 59,639  \\ \hline
Biendata  & 29,710     & 4,244        & 8,490   \\ \hline
\end{tabular}
\caption{Number of aligned complex-simple sentence pairs by dataset}
\vspace{-2mm}
\end{table}

\vspace{-2mm}
\subsection{(Psycho-)Linguistic Features} \label{cocogen}

The textual data of the three datasets were automatically analyzed using CoCoGen (short for Complexity Contour Generator), a computational tool that implements a sliding window technique to calculate sentence-level measurements for a given feature \cite[for recent applications of the tool, see][]{kerz-etal-2020-becoming,kerz2022pushing,wiechmann2022measuring}. We extracted 107 features that fall into five categories: (1) measures of syntactic complexity (N=16), (2) measures of lexical richness (N=14), (3) register-based n-gram frequency measures (N=25), (4) readability measures (N=14), and (5) psycholinguistic measures (N=38). The first category comprises (i) surface measures that concern the length of production units, such as the mean length clauses and sentences, or (ii) measures of the type and incidence of embeddings, such as dependent clauses per T-Unit or verb phrases per sentence. These features are implemented based on descriptions in \citet{lu2010automatic} using the Tregex tree pattern matching tool \cite{levy2006tregex} with syntactic parse trees for extracting specific patterns. The second category comprise several distinct sub-types, including (i) measures of lexical variation, i.e. the range of vocabulary as displayed in language use, captured by text-size corrected type-token ratio and (ii) lexical sophistication, i.e. the proportion of relatively unusual or advanced words in the learner’s text. The operationalizations of these measures follow those described in \citet{lu2012relationship} and \citet{stroebel2016}. The register-based n-gram frequency measures of the third category are derived from the five register sub-components of the Contemporary Corpus of American English \cite[COCA,][]{davies2009385+}: spoken, magazine, fiction, news and academic language \cite[see][for details]{kerz-etal-2020-becoming}. The fourth category combine a word familiarity variable defined by pre-specified vocabulary resource to estimate semantic difficulty together with a syntactic variable, such as average sentence length. Examples of these measures are the Fry index \cite{fry1968readability} or the SMOG formula \cite{mclaughlin1969clearing}. The psycholinguistic measures of the fifth category capture cognitive aspects of human language processing not directly addressed by the surface vocabulary and syntax features of traditional formulas. These measures include a word’s average age-of-acquisition \cite{kuperman2012age} or prevalence, which refers to the number of people knowing the word \cite{brysbaert2019word, johns2020estimating}. For an overview of all features, see Table \ref{tab:indicators} in the Appendix. Tokenization, sentence splitting, part-of-speech tagging, lemmatization and syntactic PCFG parsing were performed using Stanford CoreNLP \cite{manning2014stanford}.

\vspace{-2mm}
\subsection{Complexity prediction} \label{complexity prediction}
\vspace{-2mm}


For complexity prediction, i.e. the preliminary task of classifying whether a given text needs to be simplified or not, we performed experiments with five (hybrid) deep neural network architectures. Two of these prediction models are reimplementations of models used in \citet{garbacea-etal-2021-explainable} and serve as baselines: The first model, LSTM, is a 2-layer word-level BLSTM classifier that uses GloVe word embeddings as input. The second baseline model is a 12-layer BERT model for sequence classification using a pre-trained BERT, with the first 8 layers frozen during fine-tuning.


We also conducted experiments with three hybrid models that integrate the (psycho-)linguistic features described in Section \ref{cocogen} into neural networks. GloVe-PSYLING A and GloVe-PSYLING B are hybrid BLSTM with attention models \cite{maskinfill} that differ in how the integration was performed: In model A, the linguistic features were concatenated with word embeddings before being fed into a BLSTM. In the B model, the linguistic features were concatenated with the last layer hidden state of the BLSTM. In the third hybrid model, referred to as BERT-PSYLING, we concatenated the linguistic features with the last layer output for [CLS] token from BERT. The vector representation of a sentence was then fed into a MLP classifier with ReLu as activation function. For all classifiers, Best hyperparameters were found by grid search: For BERT on WiKiLarge, the best results were obtained with a learning rate of $3\times 10^{-5}$ and a batch size of 64. For LSTM on Newsela, the learning rate was $2\times10^{-4}$ and the batch size was 32. For BERT-PYSLING on Biendata, the learning rate was 2e-5 and the batch size was 32. We used Adam as the optimizer with $\beta=(0.9, 0.999)$ and $\epsilon=10^{-8}$. Early stopping, where accuracy did not increase for more than 4 epochs, was used as the stopping criterion. All models were evaluated using precision, recall, F1, and classification accuracy on balanced training, validation, and testing datasets.

\vspace{-2mm}
\subsection{Complexity Explanation} \label{complexity explanation}
\vspace{-1mm}

The objective of the complexity explanation subtask is to highlight the part of the text that needs to be simplified. In  \citet{garbacea-etal-2021-explainable} this was achieved by quantifying the relative importance of the features in the of complexity prediction models (unigrams, bigrams, trigrams, GloVe word embeddings) using model-agnostic explanatory techniques, in particular LIME \cite{ribeiro2016model} and SHAP \cite{lundberg2017unified}. To afford complexity explanation of the five complexity prediction models described in section \ref{complexity prediction}, we utilized BERT attention outputs: Since BERT uses byte-pair tokenization, we converted token attentions to word attentions by averaging the token attention weights per word. For a given attention head, the attention weights from the [CLS] token to other words at the first layer were considered as weights of those words for a given sentence. For each individual word, its final weight was the average of the weights from the 12 heads of BERT. The decision whether or not to highlight a particular word was based on a comparison of its final weight and the average of the final weights of all words in a given sentence: a word was considered complex, and thus highlighted, if its final weight fell below sentence average (see Figure \ref{fig:attention} in the Appendix). We compare these complexity explanatory approaches with LSTM-LIME, random highlighting, and lexicon-based highlighting based on words that appear in the Age-of-Acquisition (AoA) lexicon  \citet[see][for details on these basic methods]{garbacea-etal-2021-explainable}. Following \citet{garbacea-etal-2021-explainable}, we evaluated the models using token-wise precision (P), recall (R), and translation edit rate (TER) \cite{snover2006study}, which assesses the minimum number of edits needed to the unhighlighted part of a source sentence so that it exactly matches the target sentence.

\vspace{-2mm}
\subsection{Simplification Generation}
\label{simplification generation}
\vspace{-1mm}


The original AudienCe-CEntric Sentence Simplification (ACCESS) model, introduced by \citet{martin-etal-2020-controllable}, provides explicit control of TS by conditioning the output returned by the model on specific attributes. The ACCESS model used four such parameter tokens as control features: (1) NbChars, the character length ratio between source sentence and target sentence, (2) LevSim, the normalized character-level Levenshtein similarity between source and target, which quantifies the amount of modification operated on the source sentence, (3) WordRank, a proxy to lexical complexity measured as the third-quartile of log-ranks of all words in a sentence. To get a ratio the WordRank of the target was divided by that of the source. The Seq2Seq model is parametrized on the control features by prepending a these attributes and their values as additional inputs to the source sentences as plain text `parameter tokens'. The special token values are the ratio of this parameter token calculated on the target sentence with respect to its value on the source sentence. For example to control the number of characters of a generated simplification, the compression ratio between the number of characters in the source and the number of characters in the target sentence is computed. Ratios are discretized into bins of fixed width of 0.05 and capped to a maximum ratio of 2. Special tokens are then included in the vocabulary. At inference time, we the ratio is set a fixed value for all samples. For example, to generate simplifications that are 80\% of the length of the source sentence, the token <NbChars 0.8> is prepended to each source sentence. As the Seq2Seq model, a Transformer model with a base architecture \cite{Vaswani2017AttentionIA} was trained utilizing FairSeq toolkit \cite{Ott2019fairseqAF}.

Our extended model, referred to here as ACCESS-XL, integrates ten of the 107 features examined in the complexity prediction step. These ten measures were selected to cover all feature groups. Within the lexical richness group, which is the largest of the five groups, features were selected to represent all subcategories of the group, i.e. length of production unit, lexical diversity, lexical sophistication, n-gram frequency, and both crowdsourcing-based and corpus-based word prevalence. Figure \ref{fig:featureSelection} in the Appendix shows the differences in mean standardized feature scores between `normal' and `simple' sentences in Wikipedia, highlighting in blue the features selected in our model. Following \cite{martin-etal-2020-controllable}, we then trained a base transformer \cite{Vaswani2017AttentionIA} using the FairSeq toolkit \cite{fairseq}. Both encoder and decoder consist of 6 layers. For the encoder, each of the 6 layers consists of an 8-head self-attention sub-layer and a position-wise fully connected sub-layer with a  dimensionality of 2048. Each decoder layer has a similar structure, but with an additional 8-head self-attention layer that performs multi-head attention over the output of the encoder stack. The embedding size is 512. Dropout with a rate 0.2 was used for regularization. The optimizer used is the Adam optimizer with a learning rate of 0.00011, $\beta=(0.9, 0.999)$, $\epsilon=10^{-8}$. Label smoothing with a uniform prior distribution of $\epsilon=0.54$ was applied. Early stopping was used to prevent overfitting, with non-increasement of SARI score for more than 5 epochs as the stopping criterion. Sentencepiece with a vocabulary size of 10k was used as the tokenizer \cite{kudo2018sentencepiece}. Beam search with a beam size of 8 for searching for the best possible simplified sentence. A fixed combination of control tokens (a control feature along with its binned value) was used in text generation. To find the best combination, we applied the greedy forward select algorithm; we progressively added the control token from a candidate set that, in combination with the previously added control tokens, leads to the largest performance improvement in terms of SARI score on the validation set of WiKiLarge. After adding a control token to the combination, all control tokens sharing the same control feature with the newly added token were removed from the candidate set. The algorithm stopped when no control token led to an improvement in SARI score or no control token was left in the candidate set. The 5 most frequent control tokens from the WiKiLarge training set were used as the initial candidate set for each control feature, resulting in a reduction of the total search space from about $40^{10}$ to $5^{10}$. We evaluated the output of the text simplification models using the FKGL (Flesch-Kincaid Grade Level) readability metric  \cite{kincaid1975derivation} to evaluate simplicity and SARI \cite{xu2016optimizing} as an overall performance metric, since FKGL does not take into account grammaticality and meaning preservation \cite{wubben2012sentence}\footnote{See Appendix for definitions and more details on these evaluation metrics.}.
All scores were calculated using the EASSE python package for sentence simplification \cite{alva-manchego-etal-2019-easse}\footnote{\url{https://github.com/feralvam/easse}}.  We selected the model with the best SARI on the validation set and report its score on the test set. The best combination of control tokens was as follows: $\text{MLS}_{0.50}$, $\text{Fry}_{0.85}$, $\text{FORCAST}_{0.90}$, $\text{WPCorp}_{0.95}$, $\text{WPCrowd}_{0.90}$, $\text{BigramNews}_{2.00}$, $\text{ANC}_{0.85}$, $\text{AoA}_{1.00}$, $\text{MLWs}_{0.90}$, $\text{CTTR}_{0.85}$.

    

\FloatBarrier
\vspace{-2mm}
\section{Results}
\vspace{-2mm}

An overview of the results of the three subtasks -- complexity prediction, complexity explanation and simplification generation -- is presented in Table \ref{tab:results}. We discuss the results of each subtask in turn.

\begin{table*}[]

\centering
\adjustbox{max width=\textwidth}{
\begin{tabular}{|l|c|c|c|c||c|c|c|c||c|c|c|c|}
\hline
                                                            & \multicolumn{12}{c|}{\textbf{Complexity Prediction}}                                                                                        \\
                                                            & \multicolumn{4}{c}{WikiLarge}                 & \multicolumn{4}{c}{Newsela}       & \multicolumn{4}{c|}{Biendata}       \\
                                                            \hline
Model                                                       & P     & R    & F1    & Acc       & P  & R & F1 & Acc    & P  & R  & F1 & Acc    \\
\hline
LSTM \cite{garbacea-etal-2021-explainable}    & -& -     &-            &    0.716          & - &    -    &-    &0.733        &-            &-         &-    &    0.898    \\
BLSTM GloVe     &0.731&     0.710&         0.721  &     0.725          &0.867  &\textbf{0.703}        &\textbf{0.776}    &\textbf{0.797}        &0.923            &0.889         &0.906    &0.907        \\
BERT     & \textbf{0.794}  & \textbf{0.807} & \textbf{0.800} & \textbf{0.799}  & \textbf{0.973}&  0.572  & 0.720  & 0.778     &0.934            &\textbf{0.947}         &0.940    &0.940        \\
GloVe-PSYLING A (ours)    & 0.766   & 0.781   & 0.773 & 0.771     &  0.929 &  0.609 & 0.736   &  0.781 & 0.930  & 0.915  &0.922 & 0.923 \\
GloVe-PSYLING B (ours)     &  0.762   &  0.783  &  0.772   & 0.769  & 0.925           & 0.604       & 0.731   & 0.778       &0.924& 0.928 & 0.926 &  0.926      \\
BERT-PYSLING (ours)      &0.779 &  \textbf{0.807}  &    0.793 & 0.789           &0.972               &0.580         &0.727     &0.782        &\textbf{0.942}            &0.945         &\textbf{0.943}    &\textbf{0.943}        \\
\hline
                                                            & \multicolumn{12}{c|}{\textbf{Complexity Explanation}}                                                                                       \\
                                                            & \multicolumn{4}{c}{WikiLarge}                 & \multicolumn{4}{c}{Newsela}       & \multicolumn{4}{c|}{Biendata}       \\
                                     \hline                       & P     & R    & F1    & TER$\downarrow$       & P  & R & F1 & TER$\downarrow$    & P  & R  & F1 & TER$\downarrow$    \\
                                     \hline
Random highlighting                                         &  0.410              &0.463         &  0.457         &   1.084        &    0.550         &     0.488   &  0.504  &     1.029     &    0.803       &      0.424   &    0.550    &   1.011     \\
AoA lexicon                                                 &   0.407            & 0.549          &   \textbf{0.500}     &      1.026      &      0.550      &  0.620      & 0.572    &  0.858       &    0.770    & 0.629              &   0.678  &    0.989   \\
LSTM+LIME      &  0.404          &  0.639       & 0.419   & 0.997   &   0.520  &     0.615      & 0.506      &   1.062        &    0.805        &   \textbf{0.826}     &  \textbf{0.796}  &     0.983          \\
BERT                    &   0.405         &  \textbf{0.660}       & 0.434    &   \textbf{0.936}         &   0.542   &    \textbf{0.729}      &   \textbf{0.597}    & 0.817    &     0.784       &   0.635    &  0.688  &    0.965           \\
GloVe-PSYLING A (ours)                                &  \textbf{0.454}             &  0.596         & 0.426   & 1.010  &  \textbf{0.579}  &    0.481  &  0.501 &  0.827  &    0.806    &    0.552        &     0.641    &   0.959     \\
GloVe-PSYLING B (ours)         &  0.453    &  0.643       & 0.440    &  0.999         &          0.544  &    0.554    &  0.524  &   \textbf{0.816}     &    \textbf{0.813}         &  0.556         &    0.646  &    \textbf{0.951}      \\
BERT-PSYLING  (ours)           & 0.400              &  0.619         &  0.419     &   0.949        &         0.540   &   0.701     &  0.586  &     0.818   &    0.781        &    0.638     & 0.688   & 0.966   \\
\hline
                                                            & \multicolumn{12}{c|}{\textbf{Simplification Generation}}                                                                                        \\
                                                            & \multicolumn{4}{c}{WikiLarge (wd)} & \multicolumn{4}{c}{Newsela (OOD)} & \multicolumn{4}{c|}{Biendata (OOD)} \\
                                                            \hline
                                                            & \multicolumn{2}{c|}{SARI$\uparrow$}     &\multicolumn{2}{c|}{FK$\downarrow$} &
                                                            \multicolumn{2}{c|}{SARI$\uparrow$}     &\multicolumn{2}{c|}{FK$\downarrow$} &
                                                            \multicolumn{2}{c|}{SARI$\uparrow$}     &\multicolumn{2}{c|}{FK$\downarrow$} \\
                                                            \hline




ACCESS
&\multicolumn{2}{c|}{41.87}     
&\multicolumn{2}{c|}{7.22} 
&\multicolumn{2}{c|}{29.44}     
&\multicolumn{2}{c|}{6.45} 
&\multicolumn{2}{c|}{20.21}     
&\multicolumn{2}{c|}{12.53} \\ 


ACCESS-XL (ours)                   
&\multicolumn{2}{c|}{\textbf{43.34}}     
&\multicolumn{2}{c|}{\textbf{4.39}} 
&\multicolumn{2}{c|}{\textbf{34.91}}     
&\multicolumn{2}{c|}{\textbf{3.96}} 
&\multicolumn{2}{c|}{\textbf{27.25}}     
&\multicolumn{2}{c|}{\textbf{10.71}} \\ 
\hline
\end{tabular}
}
\caption{\textbf{Prediction}: Scores represent out-of-sample precision (P), recall (R), F1  and accuracy (Acc) scores. \textbf{Explanation}: P, R, and and F1 values represent token-level scores. TER scores represent Translation Edit Rates \cite{snover2006study} \textbf{Simplification}: Scores represent out-of-sample SARI and Flesch-Kinkaid Grade Level (FK)}
\label{tab:results}
\vspace{-4.5mm}
\end{table*}

\textbf{Complexity prediction:} Our best-performing models outperformed the classification accuracy of explainable model – the word-level LSTM - predicting complexity in all three benchmark datasets reported in Garbacea et al. (2021). Since the pattern of results is consistent across all evaluation metrics, we focus here on classification accuracy: On WikiLarge, we improve on the word-level LSTM presented in \citet{garbacea-etal-2021-explainable} by +8.08\% by extracting attention weights from the pre-trained BERT model. On Newsela, our GloVe-based LSTM model outperforms the word-level LSTM by +6.68\%. On the Biendata dataset, our hybrid model that integrates BERT representations with linguistic features leads to an improvement of +4.43\%. Overall, our results replicate the general pattern of results reported in \citet{garbacea-etal-2021-explainable} in that the best-performing models achieve approximately 80\% accuracy on the WikiLarge and Newsela datasets and much higher -- approximately 95\% accuracy -- on the Biendata dataset. These results support the conclusion drawn in \citet{garbacea-etal-2021-explainable} that complexity prediction is influenced by the application domain, with the distinction between scientific content and public domain press releases (Biendata) being much easier than the distinction between regular and simplified news articles (Newsela) or Wikepedia articles (WikiLarge).

 On the WikiLarge dataset, the BERT model performed the best, with a +7.4\% performance increase over the LSTM. On the Newsela dataset, however, the LSTM achieved the highest accuracy, outperforming both the BERT and BERT-HYBRID models by +1.9\% and +1.5\%, respectively. On the Biendata dataset, the highest performance was achieved by our BERT-HYBRID model, which improved the already high performance of the LSTM by +3.6\%. Across all datasets, the GloVe word embedding-based models consistently ranked between the LSTM and BERT-based models, suggesting that the use of contextualized word embeddings of the BERT-based model may reduce the generalizability of the model, leading to variations in model performance across datasets.

\textbf{Complexity explanation:} The second part of Table \ref{tab:results} presents the results of the subtask designed to evaluate how well complexity classification can be explained, as measured by how accurately the complex parts of a sentence can be identified (highlighted). In general, all of our models showed better recall than precision, meaning that they were better at identifying words that were removed in the simplified version of a pair than words that were truly removed from the complex version. This pattern is opposite to what is reported in \citet{garbacea-etal-2021-explainable}, where precision is strongly favoured over recall. This may indicate that using average attention as a threshold may not be optimal: While this approach is the de facto standard in text style transfer research, recent work has pointed out the limitations of this approach, such as its inability of handling flat attention distributions \cite{lee2021enhancing}\footnote{Figure \ref{fig:attention} in the Appendix illustrates the differences in attention weight distributions among our models.}. Future research may address this issue. As in the case of complexity prediction, we found that the performance of the models is dataset-specific and also varies with respect to the rank order across evaluation metrics: For WikiLarge, the BERT model achieved the best recall and TER scores, while precision was highest for the GloVE-based hybrid models (+4.5\% compared to BERT). For Newsela, the BERT-based models outperformed the other models in terms of recall and F1, while the GloVe-based hybrid models achieved higher precision. All of our models significantly outperformed the three base models in terms of TER values, with the best performing model, Glove-PSYLING B, reducing the TER of the AoA method by 4.2\% and that of the LSTM by as much as -24.4\%. For Biendata, Glove-PSYLING B achieved the best values for precision and TER. However, the LSTM dominated the ranking in terms of recall and F1 with improvements of the next best model (BERT-PSYLING) by up to 18\%.

\textbf{Simplification Generation} We establish the state-of-the-art at 43.34 SARI on the WikiLarge test set, an improvement of +1.47 over the best previously reported result. Our ACCESS-XL text simplification model consistently outperforms the original ACCESS model \cite{martin-etal-2020-controllable} on all datasets and performance metrics. The performance improvement was even greater in the out-of-domain settings -- with a +5.47\% increase in SARI in the Newsela dataset and +7.04\% in the Biendata dataset -- suggesting that increased controllability also leads to increased model robustness and generalizability. For FK readability, the performance gain is even more pronounced: in the within-domain setting (WikiLarge), the ACCESS-XL model achieves a Flesch-Kinkaid score of 4.39, an improvement of -2.88. To put this number in perspective, the original ACCESS model improved previous state-of-the-art models, SBMT+PPDB+SARI \cite{xu2016optimizing} and PBMT-R \cite{wubben2012sentence}, by only -0.07 and -1.11, respectively. As in the case of SARI, the improvement in FK performance extends to both out-of-domain settings with an improvement of -2.49 for Newsela and -1.82 for Biendata. To shed more light on the textual characteristics of the outputs of the two text simplification models, we compared their average scores on the ten parameter tokens. A visualization of the results along with the scores obtained for the target and source sentences of the testset for each dataset is shown in Figure \ref{fig:behavior} in the Appendix. The comparisons revealed several important facts about the behavior of the models as well as the training data: (1) For the WikiLarge dataset, on which the model was trained, we found that the differences in average scores between the `complex' source sentences and the `simple' target sentences varied in magnitude: On some measures, such as mean sentence length (MLS) -- a proxy of syntactic complexity, the difference between simple and complex sentences is very pronounced (MLS\textsubscript{simple}=14.9 words, MLS\textsubscript{complex}=22.4 words). For others, e.g. LS.ANC -- a measure of lexical sophistication, the difference between the standard versions and their simplified counterparts is minimal (LS.ANC\textsubscript{simple}=0.411, LS.ANC\textsubscript{complex}=0.414). These results are consistent with previous indications of limitations in the WikiLarge dataset related to the high proportion of inappropriate simplifications \cite{xu2016optimizing}. We further observed (2) that the ACCESS-XL model successfully learned to control the attributes and achieved the desired effect on the generated simplifications: For example, its outputs are characterized by much lower MLS values (MLS\textsubscript{ACCESS-XL} = 10.8 words) compared to the source. We note that shorter MLS values were achieved by splitting the sentence (rather than simply deleting content), which has been shown to be a weakness of current seq2seq TS models \cite{maddela2020controllable}. This is illustrated in the sentence set in Table \ref{tab:example} in the Appendix. And (3) we found that the ACCESS-XL model was able to successfully generalize its ability to control the target attributes to out-of-domain settings. For example, the learned control over the MLS parameter led to the generation of Newsela simplifications that almost matched almost perfectly the mean value of the simple sentence targets in this dataset. 

\vspace{-1.8mm}
Lastly, we address the question of whether explainable prediction of text complexity is still a necessary preliminary step in the pipeline when using a strong, end-to-end simplification system. We found that for all datasets -- and for both the original ACCESS model and the extended ACCESS-XL model -- using of preliminary complexity prediction did not improve simplification performance (see Figure \ref{fig:sari} in the Appendix): For both SARI and FKGL evaluation metrics the best performance was invariably achieved by a model without prior indication of what sentences should undergo simplification. These results stand in stark contrast to the results reported in \citet{garbacea-etal-2021-explainable}, where prior complexity prediction was found to improve the performance of the original ACCESS model. Rather than evaluating performance using SARI and FKGL, as was the case here and in the original ACCESS publication \cite{martin-etal-2020-controllable}, \citet{garbacea-etal-2021-explainable} evaluated model performance using edit distance (ED), TER, and Frechet Embedding Distance. For ED alone, the reported improvements ranged from 30\% to 50\%. Follow up experiments based on ED, conducted to determine if the discrepancy was related to the choice of evaluation metric only confirmed the pattern of results reported here for SARI and FKGL (see Tables \ref{evalViaED} and \ref{evalViaEDcomplex} in the Appendix). Follow-up experiments based on ED, conducted to determine if the discrepancy was related to the choice of scoring metric, only confirmed the pattern of results reported here for SARI and FKGL (see Tables \ref{evalViaED} and \ref{evalViaEDcomplex} in the Appendix). \citet{garbacea-etal-2021-explainable} conclude that the ACCESS model -- and also the DMLMTL presented in \cite{guo2018dynamic}, which had the highest performance for Newsela (33.22 SARI) -- tends to simplify even simple inputs. Moreover, \cite{garbacea-etal-2021-explainable} report that over 70\% of the `simple' sentences in the test data were modified (and thus oversimplified) by the ACCESS model. Note, however, that `simple' here means that the input sentence in question was classified as such by a preliminary complexity prediction model. Since these classifiers in WikiLarge only achieve a classification accuracy of 80\%, the true percentage of oversimplification cannot be accurately estimated.

\vspace{-2mm}
\section{Conclusion and Future Work}
\vspace{-2mm}

In this work, we have advanced research on explainable and controllable text simplification in two ways: First, we have shown that performance on a prior task of explainable complexity prediction can be significantly improved by the combined use of (psycho-)linguistic features and pre-trained neural language models. And second, by extending the AudienCe-CEntric sentence simplification model to explicitly control ten text attributes, we have achieved a new state of the art in text simplification in both within-domain and out-of domain settings. In future work, we plan to apply our modeling approach to another key text style transfer task, that of formality transfer, and evaluate it on existing benchmark datasets such as the GYAFC dataset \cite{rao2018dear}. Moreover, we intend to explore the role of (psycho-)linguistic features for controllable TS in unsupervised settings using a variational auto-encoder and a content predictor in combination with attribute predictors \cite{liu2020revision}.

\section{Limitations}


The current work relies exclusively on automatic evaluation metrics for text simplification. 
While such metrics provide a cost-effective, reproducible, and scalable way to gauge the quality of text generation results, they also have their own weaknesses. Human scoring is necessary to address some of the inherent weaknesses of automatic evaluation \cite[for more details, see][]{jin2022deep}

Furthermore, the performance of the proposed text simplification methods was tested on informational texts in English. While we assume that the methods can be applied to other domains and languages, we have not tested this assumption experimentally and limit our conclusions to English and the types of language registers represented in the three datasets used in this work.




\bibliography{custom}
\bibliographystyle{acl_natbib}

\clearpage
\newpage
\FloatBarrier

\onecolumn
\section{Appendix} \label{sec:Appendix}

\begin{table*}[h!]
  \centering
  \setlength{\tabcolsep}{2pt}
  \caption{Overview of the 107 features investigated in the work}
    \begin{tabular}{|l|c|l|l|}
		\hline
		Feature group & Number & Features & Example/Description \\
		& of features &  & \\
		\hline
		Syntactic complexity & 16    &MLC&Mean length of clause (words)\\

		&&MLS&Mean length of sentence (words)\\
		&&MLT&Mean length of T-unit (words)\\
		&&C/S&Clauses per sentence\\
		&&C/T&Clauses per T-unit\\
		&&DepC/C&Dependent clauses per clause\\
		&&T/S&T-units per sentence\\
		&&CompT/T&Complex T-unit per T-unit\\
		&&DepC/T&Dependent Clause per T-unit\\
		&&CoordP/C&Coordinate phrases per clause\\
		&&CoordP/T&Coordinate phrases per T-unit\\
		&&NP.PostMod&NP post-mod (word)\\
		&&NP.PreMod&NP pre-mod (word)\\
		&&CompN/C&Complex nominals per clause\\
		&&CompN/T&Complex nominals per T-unit\\
		&&VP/T&Verb phrases per T-unit\\
		\hline
		Lexical richness & 14    &MLWc& Mean length per word  (characters)\\
		&&MLWs&Mean length per word (sylables)\\
		&&LD&Lexical density\\
		&&NDW&Number of different words\\
		&&CNDW&NDW corrected by Number of words\\
		&&TTR&Type-Token Ration (TTR)\\
		&&cTTR&Corrected TTR\\
		&&rTTR&Root TTR\\
		&&AFL&Sequences Academic Formula List\\
		&&ANC& LS (ANC) (top 2000, inverted)\\
		&&BNC&LS (BNC) (top 2000, inverted)\\
		&&NAWL&LS New Academic Word List\\
		&&NGSL&LS (General Service List) (inverted)\\
		&&NonStopWordsRate& Ratio of words in NLTK non-stopword list\\
		\hline
		Register-based  & 25    & Spoken ($n\in [1,5]$) & Frequencies of uni-, bi-\\
		&       & Fiction ($n\in [1,5]$) & tri-, four-, five-grams\\
		&       & Magazine ($n\in [1,5]$) &  from the five sub-components \\
		&       & News ($n\in [1,5]$) & (genres) of the COCA\\
		&       & Academic ($n\in [1,5]$) &  \\
		\hline
	\end{tabular}%
  \label{tab:indicators}%
\end{table*}

\twocolumn
\begin{table*}
  \centering
  \setlength{\tabcolsep}{1pt}
    \begin{tabular}{|l|c|l|l|}
		\hline
		Feature group & Number & Features & Example/Description \\
		& of features &  & \\
		\hline
		Readability & 14     & ARI  &  Automated Readability Index \\
		&       & ColemanLiau & Coleman-Liau Index \\
		&       &DaleChall &  Dale-Chall readability score\\
		&&FleshKincaidGradeLevel&Flesch-Kincaid Grade Level\\ 
		&&FleshKincaidReadingEase&Flesch Reading Ease score\\
		&&Fry-x&x coord. on Fry Readability Graph\\
		&&Fry-y&y coord. on Fry Readability Graph\\
		&&Lix&Lix readability score\\
		&&SMOG&Simple Measure of Gobbledygook\\
		&&GunningFog&Gunning Fog Index readability score\\
		&&DaleChallPSK&Powers-Sumner-Kearl Variation of \\
		&&&the Dale and Chall Readability score\\
		&&FORCAST&FORCAST readability score\\
		&&Rix&Rix readability score\\
		&&Spache&Spache readability score\\
		
		\hline
		Psycholinguistic  & 38    &  WordPrevalence & See \citet{brysbaert2019word} \\
		&& Prevalence & Word prevalence list\\
		&&&incl. 35 categories \\
		&&&(\citet{johns2020estimating})\\
		&&AoA-mean&avg. age of acquisition \\
		&&&(\citet{kuperman2012age})\\
		&&AoA-max&max. age of acquisition\\
		\hline
	\end{tabular}%
\end{table*}

\begin{table*}[]
  \caption{Means and standard deviations of all engineered languge features across the `normal' and `simple' sentences in the three benchmark datasets}
\centering
\setlength{\tabcolsep}{1pt}
\begin{tabular}{|l|cc|cc||cc|cc||cc|cc|}
\hline
                        & \multicolumn{4}{c||}{Biendata}                            & \multicolumn{4}{c||}{Newsela}                             & \multicolumn{4}{c|}{WikiLarge}                           \\
                        & \multicolumn{2}{c}{normal} & \multicolumn{2}{c||}{simple} & \multicolumn{2}{c|}{normal} & \multicolumn{2}{c||}{simple} & \multicolumn{2}{c|}{normal} & \multicolumn{2}{c|}{simple} \\
\hline
Feature                 & M            & SD          & M            & SD          & M            & SD          & M            & SD          & M            & SD          & M            & SD          \\
\hline
LexDens                 & 0.73         & 0.1         & 0.76         & 0.12        & 0.58         & 0.1         & 0.58         & 0.11        & 0.58         & 0.12        & 0.6          & 0.17        \\
CTTR                    & 3.9          & 0.66        & 3.53         & 0.56        & 4.69         & 0.83        & 3.94         & 0.67        & 4.33         & 0.92        & 3.71         & 1.06        \\
RTTR                    & 2.65         & 0.42        & 2.39         & 0.37        & 3.21         & 0.57        & 2.69         & 0.44        & 2.96         & 0.63        & 2.54         & 0.7         \\
TTR                     & 0.97         & 0.05        & 0.99         & 0.04        & 0.91         & 0.07        & 0.95         & 0.06        & 0.88         & 0.1         & 0.92         & 0.1         \\
MLWc                    & 6.59         & 1.09        & 5.87         & 1.03        & 4.89         & 0.61        & 4.67         & 0.67        & 4.98         & 0.83        & 4.95         & 1.19        \\
MLWs                    & 2.02         & 0.36        & 1.73         & 0.35        & 1.47         & 0.2         & 1.39         & 0.21        & 1.52         & 0.25        & 1.49         & 0.37        \\
Prev.AllAP              & 6.25         & 1.07        & 7.12         & 0.74        & 7.3          & 0.59        & 7.38         & 0.68        & 6.54         & 1.25        & 6.51         & 1.5         \\
Prev.AllBP              & 7.48         & 1.31        & 8.58         & 0.95        & 8.98         & 0.75        & 9.11         & 0.87        & 8            & 1.56        & 7.96         & 1.86        \\
Prev.AllCD              & 9.43         & 1.73        & 10.65        & 1.4         & 11.98        & 1.13        & 12.25        & 1.3         & 10.66        & 2.14        & 10.6         & 2.59        \\
Prev.AllSD              & 7.55         & 1.35        & 8.79         & 1           & 9.14         & 0.77        & 9.32         & 0.89        & 8.14         & 1.57        & 8.14         & 1.89        \\
Prev.AllSDAP            & 3.63         & 0.69        & 4.23         & 0.5         & 4.44         & 0.38        & 4.51         & 0.44        & 3.95         & 0.77        & 3.93         & 0.93        \\
Prev.AllSDBP            & 5.06         & 0.98        & 5.91         & 0.75        & 6.34         & 0.57        & 6.47         & 0.66        & 5.61         & 1.12        & 5.59         & 1.36        \\
Prev.AllWF              & 10.03        & 1.85        & 11.18        & 1.51        & 12.74        & 1.22        & 13.01        & 1.41        & 11.39        & 2.3         & 11.31        & 2.79        \\
Prev.FemAP           & 5.58         & 1.02        & 6.45         & 0.71        & 6.67         & 0.55        & 6.75         & 0.64        & 5.95         & 1.15        & 5.93         & 1.38        \\
Prev.FemBP           & 6.72         & 1.26        & 7.81         & 0.92        & 8.26         & 0.71        & 8.4          & 0.83        & 7.32         & 1.45        & 7.3          & 1.74        \\
Prev.FemCD           & 8.79         & 1.69        & 9.98         & 1.39        & 11.37        & 1.1         & 11.65        & 1.27        & 10.09        & 2.05        & 10.04        & 2.49        \\
Prev.FemSD           & 6.96         & 1.31        & 8.18         & 0.99        & 8.6          & 0.74        & 8.79         & 0.86        & 7.64         & 1.49        & 7.64         & 1.8         \\
Prev.FemSDAP         & 3.01         & 0.62        & 3.56         & 0.46        & 3.79         & 0.34        & 3.86         & 0.39        & 3.34         & 0.67        & 3.33         & 0.81        \\
Prev.FemSDBP         & 4.35         & 0.91        & 5.16         & 0.72        & 5.63         & 0.53        & 5.76         & 0.61        & 4.94         & 1.02        & 4.93         & 1.24        \\
Prev.FemWF           & 9.2          & 1.78        & 10.32        & 1.48        & 11.91        & 1.18        & 12.19        & 1.36        & 10.62        & 2.18        & 10.55        & 2.66        \\
Prev.MaleAP             & 5.69         & 0.97        & 6.47         & 0.67        & 6.63         & 0.53        & 6.7          & 0.62        & 5.95         & 1.13        & 5.92         & 1.36        \\
Prev.MaleBP             & 6.99         & 1.23        & 8.01         & 0.89        & 8.38         & 0.7         & 8.51         & 0.81        & 7.48         & 1.45        & 7.45         & 1.74        \\
Prev.MaleCD             & 9.01         & 1.67        & 10.18        & 1.36        & 11.5         & 1.09        & 11.76        & 1.26        & 10.23        & 2.06        & 10.18        & 2.5         \\
Prev.MaleSD             & 7.23         & 1.3         & 8.41         & 0.97        & 8.79         & 0.74        & 8.96         & 0.86        & 7.82         & 1.51        & 7.82         & 1.82        \\
Prev.MaleSDAP           & 2.92         & 0.56        & 3.39         & 0.4         & 3.57         & 0.3         & 3.62         & 0.35        & 3.18         & 0.62        & 3.16         & 0.75        \\
Prev.MaleSDBP           & 4.45         & 0.87        & 5.18         & 0.66        & 5.59         & 0.51        & 5.7          & 0.58        & 4.95         & 0.99        & 4.93         & 1.2         \\
Prev.MaleWF             & 9.48         & 1.78        & 10.56        & 1.46        & 12.11        & 1.18        & 12.37        & 1.36        & 10.84        & 2.2         & 10.75        & 2.68        \\
Prev.UKAP               & 4.97         & 0.9         & 5.73         & 0.63        & 5.93         & 0.49        & 6            & 0.56        & 5.31         & 1.02        & 5.29         & 1.23        \\
Prev.UKBP               & 6.22         & 1.16        & 7.2          & 0.85        & 7.61         & 0.66        & 7.73         & 0.76        & 6.78         & 1.33        & 6.75         & 1.6         \\
Prev.UKCD               & 8.26         & 1.59        & 9.38         & 1.33        & 10.72        & 1.05        & 10.99        & 1.21        & 9.52         & 1.94        & 9.47         & 2.36        \\
Prev.UKSD               & 6.46         & 1.22        & 7.6          & 0.93        & 7.97         & 0.69        & 8.15         & 0.81        & 7.07         & 1.38        & 7.08         & 1.67        \\
Prev.UKSDAP             & 2.42         & 0.5         & 2.85         & 0.38        & 3.05         & 0.28        & 3.1          & 0.32        & 2.71         & 0.54        & 2.7          & 0.66        \\
Prev.UKSDBP             & 3.79         & 0.79        & 4.47         & 0.63        & 4.89         & 0.47        & 5.01         & 0.54        & 4.32         & 0.89        & 4.31         & 1.08        \\
Prev.UKWF               & 8.72         & 1.7         & 9.75         & 1.43        & 11.33        & 1.14        & 11.59        & 1.31        & 10.11        & 2.09        & 10.03        & 2.55        \\
Prev.USAAP              & 5.84         & 1.01        & 6.68         & 0.7         & 6.86         & 0.55        & 6.94         & 0.64        & 6.14         & 1.18        & 6.11         & 1.41        \\
Prev.USABP              & 7.08         & 1.27        & 8.15         & 0.92        & 8.56         & 0.72        & 8.69         & 0.84        & 7.61         & 1.49        & 7.58         & 1.78        \\
Prev.USACD              & 9.12         & 1.7         & 10.33        & 1.39        & 11.67        & 1.11        & 11.95        & 1.29        & 10.38        & 2.09        & 10.33        & 2.54        \\
Prev.USASD              & 7.25         & 1.33        & 8.49         & 0.99        & 8.86         & 0.75        & 9.04         & 0.88        & 7.87         & 1.52        & 7.88         & 1.84        \\
Prev.USASDAP            & 3.24         & 0.63        & 3.8          & 0.46        & 4.01         & 0.35        & 4.07         & 0.4         & 3.54         & 0.7         & 3.53         & 0.85        \\
Prev.USASDBP            & 4.67         & 0.93        & 5.49         & 0.72        & 5.94         & 0.54        & 6.06         & 0.63        & 5.23         & 1.06        & 5.21         & 1.28        \\
Prev.USAWF              & 9.55         & 1.81        & 10.68        & 1.48        & 12.24        & 1.19        & 12.51        & 1.38        & 10.93        & 2.23        & 10.85        & 2.71        \\
AFL                     & 0            & 0           & 0            & 0.01        & 0            & 0.01        & 0            & 0.01        & 0            & 0.01        & 0            & 0.01        \\
ANC                     & 0.53         & 0.15        & 0.46         & 0.17        & 0.32         & 0.12        & 0.29         & 0.14        & 0.42         & 0.16        & 0.42         & 0.22        \\
BNC                     & 0.7          & 0.12        & 0.67         & 0.14        & 0.53         & 0.11        & 0.51         & 0.14        & 0.6          & 0.14        & 0.62         & 0.18        \\
NAWL                    & 0.07         & 0.08        & 0.05         & 0.08        & 0.01         & 0.03        & 0.01         & 0.03        & 0.02         & 0.04        & 0.01         & 0.05        \\
NGSL                    & 0.43         & 0.16        & 0.29         & 0.16        & 0.22         & 0.12        & 0.19         & 0.13        & 0.35         & 0.18        & 0.35         & 0.23        \\

\hline
\end{tabular}
\label{featuresStats}
\end{table*}

\begin{table*}[]
\setlength{\tabcolsep}{1pt}
\begin{tabular}{|l|cc|cc||cc|cc||cc|cc|}
\hline
                        & \multicolumn{4}{c||}{Biendata}                            & \multicolumn{4}{c||}{Newsela}                             & \multicolumn{4}{c|}{WikiLarge}                           \\
                        & \multicolumn{2}{c}{normal} & \multicolumn{2}{c||}{simple} & \multicolumn{2}{c|}{normal} & \multicolumn{2}{c||}{simple} & \multicolumn{2}{c|}{normal} & \multicolumn{2}{c|}{simple} \\
\hline
Feature                 & M            & SD          & M            & SD          & M            & SD          & M            & SD          & M            & SD          & M            & SD          \\
\hline
ngram1acad          & 100.3        & 45.99       & 82.6         & 30.9        & 218.21       & 97.56       & 134.33       & 54.24       & 191.58       & 106.77      & 134.35       & 89.59       \\
ngram1fic           & 80.26        & 39.83       & 73.47        & 29.2        & 211.26       & 93.65       & 132.17       & 53.26       & 179.99       & 101.04      & 127.55       & 85.76       \\
ngram1mag           & 94.13        & 43.86       & 82.86        & 30.58       & 222.63       & 98.35       & 137.77       & 54.8        & 191.85       & 106.55      & 135.32       & 89.95       \\
ngram1news          & 86.11        & 43.24       & 78.63        & 30.42       & 222.82       & 98.5        & 137.91       & 54.84       & 190.63       & 105.75      & 134.58       & 89.44       \\
ngram1spok          & 84.23        & 42.43       & 77.21        & 30.28       & 218.49       & 97.09       & 136.38       & 54.85       & 183.53       & 103.33      & 130.18       & 87.59       \\
ngram2acad          & 11.07        & 12.58       & 8.13         & 9.15        & 41.46        & 30.02       & 27.72        & 21.32       & 32.37        & 28.85       & 24.59        & 24.74       \\
ngram2fic           & 3.55         & 5.37        & 4.26         & 6.69        & 33.83        & 25.95       & 25.47        & 20.46       & 22.31        & 21.55       & 18.68        & 19.98       \\
ngram2mag           & 7.87         & 9.44        & 8.24         & 9.36        & 45.95        & 31.26       & 31.95        & 22.9        & 32.18        & 27.69       & 25.37        & 24.67       \\
ngram2news          & 6.25         & 8.25        & 6.35         & 8.02        & 47.49        & 32.39       & 32.88        & 23.57       & 31.52        & 27.44       & 24.99        & 24.53       \\
ngram2spok          & 5.45         & 7.44        & 6.04         & 7.99        & 42.87        & 31          & 31.11        & 23.43       & 26.57        & 24.46       & 22.08        & 22.77       \\
ngram3acad          & 0.82         & 1.97        & 0.56         & 1.5         & 3.81         & 5.23        & 2.89         & 4.53        & 3.12         & 5.06        & 2.72         & 4.67        \\
ngram3fic           & 0.15         & 0.65        & 0.24         & 0.96        & 2.58         & 4.17        & 2.37         & 4.07        & 1.4          & 2.68        & 1.44         & 2.88        \\
ngram3mag           & 0.47         & 1.3         & 0.58         & 1.57        & 4.52         & 5.8         & 3.65         & 5.25        & 2.87         & 4.53        & 2.67         & 4.41        \\
ngram3news          & 0.36         & 1.12        & 0.42         & 1.26        & 4.91         & 6.19        & 3.96         & 5.61        & 2.85         & 4.62        & 2.68         & 4.53        \\
ngram3spok          & 0.28         & 1           & 0.36         & 1.22        & 3.87         & 5.68        & 3.41         & 5.33        & 1.95         & 3.58        & 2.05         & 3.86        \\
ngram4acad          & 0.09         & 0.42        & 0.06         & 0.32        & 0.41         & 1.06        & 0.34         & 1.02        & 0.35         & 1.04        & 0.32         & 0.97        \\
ngram4fic           & 0.01         & 0.13        & 0.02         & 0.2         & 0.24         & 0.76        & 0.24         & 0.82        & 0.12         & 0.42        & 0.13         & 0.5         \\
ngram4mag           & 0.05         & 0.26        & 0.07         & 0.35        & 0.52         & 1.21        & 0.45         & 1.21        & 0.31         & 0.89        & 0.31         & 0.91        \\
ngram4news          & 0.04         & 0.23        & 0.04         & 0.26        & 0.57         & 1.29        & 0.5          & 1.28        & 0.3          & 0.92        & 0.29         & 0.94        \\
ngram4spok          & 0.03         & 0.19        & 0.04         & 0.25        & 0.41         & 1.13        & 0.4          & 1.17        & 0.19         & 0.69        & 0.21         & 0.79        \\
ngram5acad          & 0.01         & 0.16        & 0.01         & 0.09        & 0.07         & 0.33        & 0.05         & 0.35        & 0.05         & 0.3         & 0.05         & 0.29        \\
ngram5fic           & 0            & 0.03        & 0            & 0.06        & 0.03         & 0.18        & 0.03         & 0.2         & 0.01         & 0.1         & 0.02         & 0.14        \\
ngram5mag           & 0.01         & 0.09        & 0.01         & 0.1         & 0.09         & 0.38        & 0.07         & 0.38        & 0.05         & 0.25        & 0.04         & 0.24        \\
ngram5news          & 0            & 0.07        & 0.01         & 0.08        & 0.09         & 0.37        & 0.08         & 0.38        & 0.05         & 0.28        & 0.04         & 0.28        \\
ngram5spok          & 0            & 0.06        & 0            & 0.07        & 0.06         & 0.31        & 0.06         & 0.32        & 0.03         & 0.18        & 0.03         & 0.21        \\
NonStopW        & 0.74         & 0.1         & 0.78         & 0.12        & 0.6          & 0.1         & 0.59         & 0.12        & 0.63         & 0.12        & 0.64         & 0.17        \\
AoA.max                 & 12.64        & 2.47        & 10.89        & 2.53        & 10.19        & 2.33        & 8.4          & 2.16        & 10.36        & 2.78        & 8.96         & 3.25        \\
AoA.mean                & 7.43         & 1.34        & 6.8          & 1.31        & 5.55         & 0.72        & 5.22         & 0.74        & 5.73         & 1.16        & 5.45         & 1.68        \\
WordPrev                & 1.62         & 0.42        & 2.04         & 0.29        & 1.99         & 0.28        & 2.01         & 0.33        & 1.62         & 0.49        & 1.59         & 0.58        \\
KolDef                  & 0.85         & 0.12        & 0.93         & 0.12        & 0.77         & 0.12        & 0.89         & 0.13        & 0.8          & 0.23        & 0.93         & 0.35        \\
NPPostMod             & 6.41         & 5.6         & 2.8          & 3.3         & 3.99         & 5.76        & 2.03         & 3.17        & 5.64         & 6.44        & 3.58         & 4.73        \\
NPPreMod              & 1.27         & 1.14        & 1.02         & 0.88        & 1.03         & 0.86        & 0.91         & 0.73        & 1.21         & 1.01        & 1.04         & 0.87        \\
CpS                     & 0.31         & 0.5         & 0.77         & 0.69        & 2.11         & 1.23        & 1.58         & 0.86        & 1.45         & 1.01        & 1.19         & 0.93        \\
CpT                     & 0.27         & 0.47        & 0.66         & 0.66        & 1.88         & 1.07        & 1.49         & 0.8         & 1.28         & 0.8         & 1.08         & 0.77        \\
CompNompC               & 0.67         & 1.23        & 1.01         & 1.08        & 1.57         & 1.2         & 1.09         & 0.9         & 1.97         & 1.51        & 1.3          & 1.24        \\
CompNompT               & 0.8          & 1.29        & 1.15         & 1.13        & 2.65         & 1.85        & 1.52         & 1.18        & 2.5          & 1.85        & 1.61         & 1.52        \\
CompTpT                 & 0.02         & 0.13        & 0.1          & 0.3         & 0.53         & 0.49        & 0.37         & 0.48        & 0.27         & 0.44        & 0.2          & 0.39        \\
CoordPpC                & 0.12         & 0.36        & 0.06         & 0.24        & 0.32         & 0.52        & 0.18         & 0.39        & 0.45         & 0.68        & 0.28         & 0.52        \\
CoordPpT                & 0.15         & 0.41        & 0.07         & 0.26        & 0.51         & 0.72        & 0.23         & 0.47        & 0.56         & 0.79        & 0.34         & 0.61        \\
DCpC                    & 0.04         & 0.17        & 0.12         & 0.29        & 0.3          & 0.29        & 0.2          & 0.27        & 0.16         & 0.27        & 0.12         & 0.24        \\
DCpT                    & 0.02         & 0.14        & 0.1          & 0.32        & 0.77         & 0.9         & 0.45         & 0.66        & 0.34         & 0.62        & 0.24         & 0.53        \\
MLC                     & 3.71         & 6.28        & 5.83         & 4.96        & 12.23        & 6.89        & 9.4          & 4.45        & 14.63        & 8.87        & 10.5         & 7.5         \\
MLS                     & 12.76        & 4.46        & 9.62         & 2.86        & 22.76        & 9.77        & 13.74        & 5.12        & 21.13        & 10.6        & 14.67        & 9.03        \\
MLT                     & 4.62         & 6.69        & 6.67         & 5.02        & 20.54        & 10.24       & 12.96        & 5.47        & 18.77        & 11.09       & 12.95        & 9.35        \\
TpS                     & 0.36         & 0.48        & 0.7          & 0.48        & 1.06         & 0.39        & 1            & 0.27        & 0.99         & 0.44        & 0.88         & 0.47        \\
VPpT                    & 0.41         & 0.64        & 0.93         & 0.82        & 2.46         & 1.42        & 1.87         & 1.04        & 1.56         & 1.08        & 1.26         & 0.98        \\
\hline
\end{tabular}
\end{table*}

\begin{table*}[]
\setlength{\tabcolsep}{1pt}
\begin{tabular}{|l|cc|cc||cc|cc||cc|cc|}
\hline
                        & \multicolumn{4}{c||}{Biendata}                            & \multicolumn{4}{c||}{Newsela}                             & \multicolumn{4}{c|}{WikiLarge}                           \\
                        & \multicolumn{2}{c}{normal} & \multicolumn{2}{c||}{simple} & \multicolumn{2}{c|}{normal} & \multicolumn{2}{c||}{simple} & \multicolumn{2}{c|}{normal} & \multicolumn{2}{c|}{simple} \\
\hline
Feature                 & M            & SD          & M            & SD          & M            & SD          & M            & SD          & M            & SD          & M            & SD          \\
\hline
ARI                     & 15.98        & 5.01        & 11.02        & 4.62        & 12.98        & 5.66        & 7.45         & 3.85        & 12.63        & 6.14        & 9.27         & 6.19        \\
Coleman             & 54.6         & 26.05       & 35.59        & 16.63       & 111.87       & 57.51       & 58.53        & 30.06       & 102.55       & 62.16       & 64.31        & 52.94       \\
DaleChall               & 10.16        & 2.13        & 8.9          & 2.7         & 6.2          & 1.96        & 5.44         & 2.3         & 7.58         & 2.49        & 7.48         & 3.37        \\
DC.PSK            & 11.41        & 1.53        & 10.31        & 1.95        & 9.06         & 1.53        & 8.01         & 1.68        & 9.99         & 1.8         & 9.56         & 2.37        \\
FK Grade  & 13.23        & 4.27        & 8.53         & 4.05        & 10.61        & 4.55        & 6.16         & 3.15        & 10.56        & 4.97        & 7.72         & 5.16        \\
FK Read & 22.95        & 30.03       & 51.06        & 29.3        & 59.52        & 19.89       & 75.39        & 18.56       & 56.99        & 23.43       & 65.85        & 31.32       \\
FORCAST                 & 13.23        & 2.14        & 11.86        & 2.62        & 9.79         & 1.67        & 9.23         & 1.96        & 10.2         & 1.93        & 10.08        & 2.91        \\
Fry.x                   & 202.05       & 36.22       & 172.58       & 35.25       & 146.84       & 19.97       & 138.88       & 21.31       & 151.77       & 25.29       & 149.05       & 37.25       \\
Gunning              & 510.4       & 178.2      & 385.0       & 114.4      & 910.4       & 390.9      & 549.6       & 204.7      & 846.5       & 423.0      & 587.4       & 361.1      \\
Lix                     & 61.4         & 14.53       & 48.27        & 16.85       & 48.26        & 14.61       & 35.23        & 12.96       & 48.96        & 15.67       & 41.45        & 19.55       \\
Rix                     & 5.9          & 2.98        & 5.33         & 2.51        & 15.49        & 6.96        & 9.91         & 4.12        & 13.96        & 7.55        & 10.08        & 6.6         \\
SMOG                    & 8.78         & 1.64        & 7.18         & 2.18        & 6.26         & 1.55        & 5.49         & 1.86        & 6.45         & 1.71        & 5.88         & 2.23        \\
Spache                  & 2.25         & 0.54        & 1.86         & 0.34        & 3.41         & 1.17        & 2.33         & 0.61        & 3.24         & 1.27        & 2.47         & 1.08 \\
\hline
\end{tabular}
\end{table*}

\FloatBarrier
\newpage

\begin{figure*}
    \centering
    \includegraphics[width = 1\textwidth]{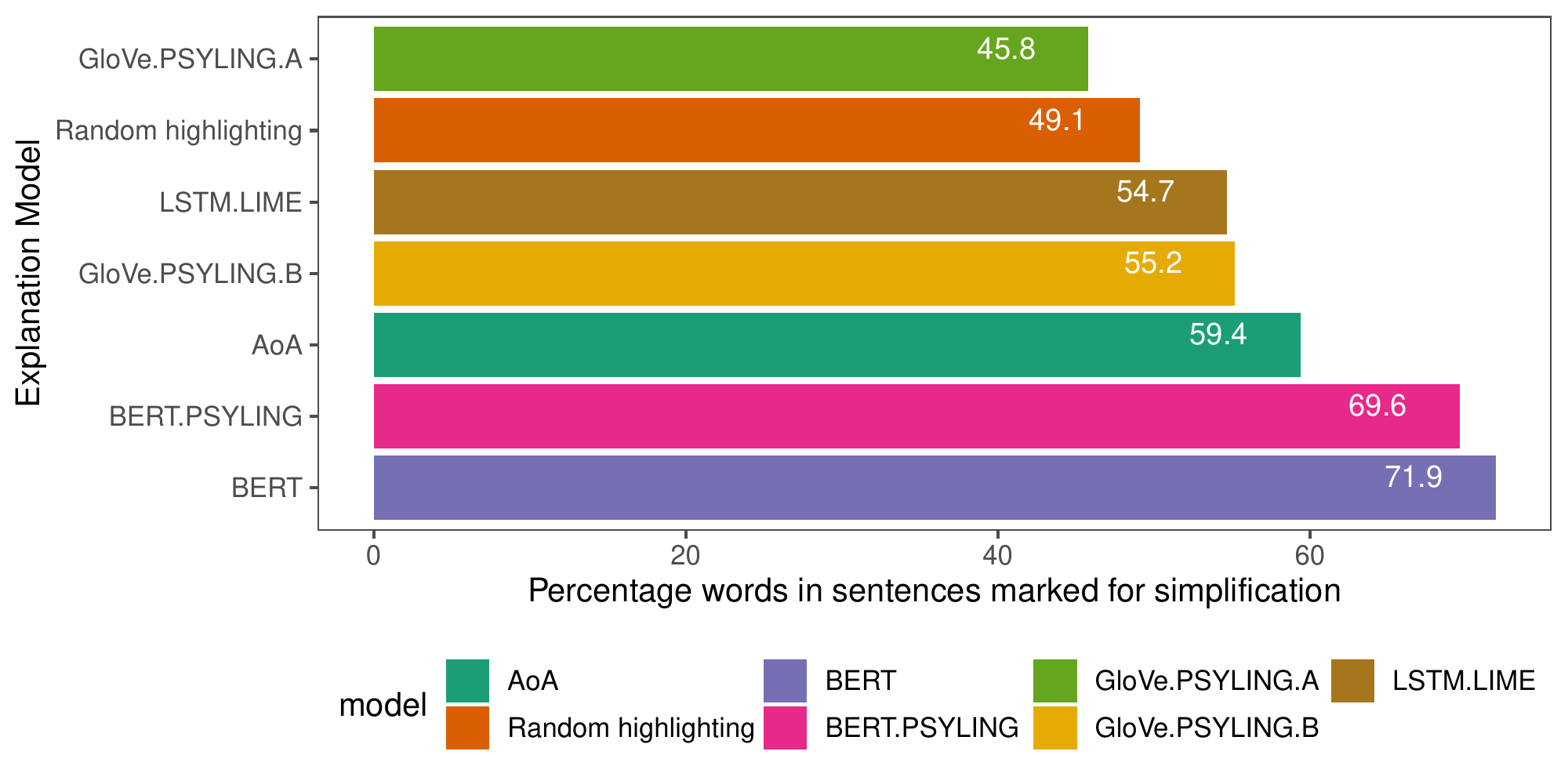}
    \caption{\textbf{Complexity explanation:} Differences in mean percentages of highlighted words across the five explanation models compared along with the two baselines: 'Random highlighting' and highlighting based on AoA (=age of acquisition) lexicon \cite{kuperman2012age}. }
    \label{fig:genPropChange}
\end{figure*}

\begin{figure*}
    \centering
    \includegraphics[width = 1\textwidth]{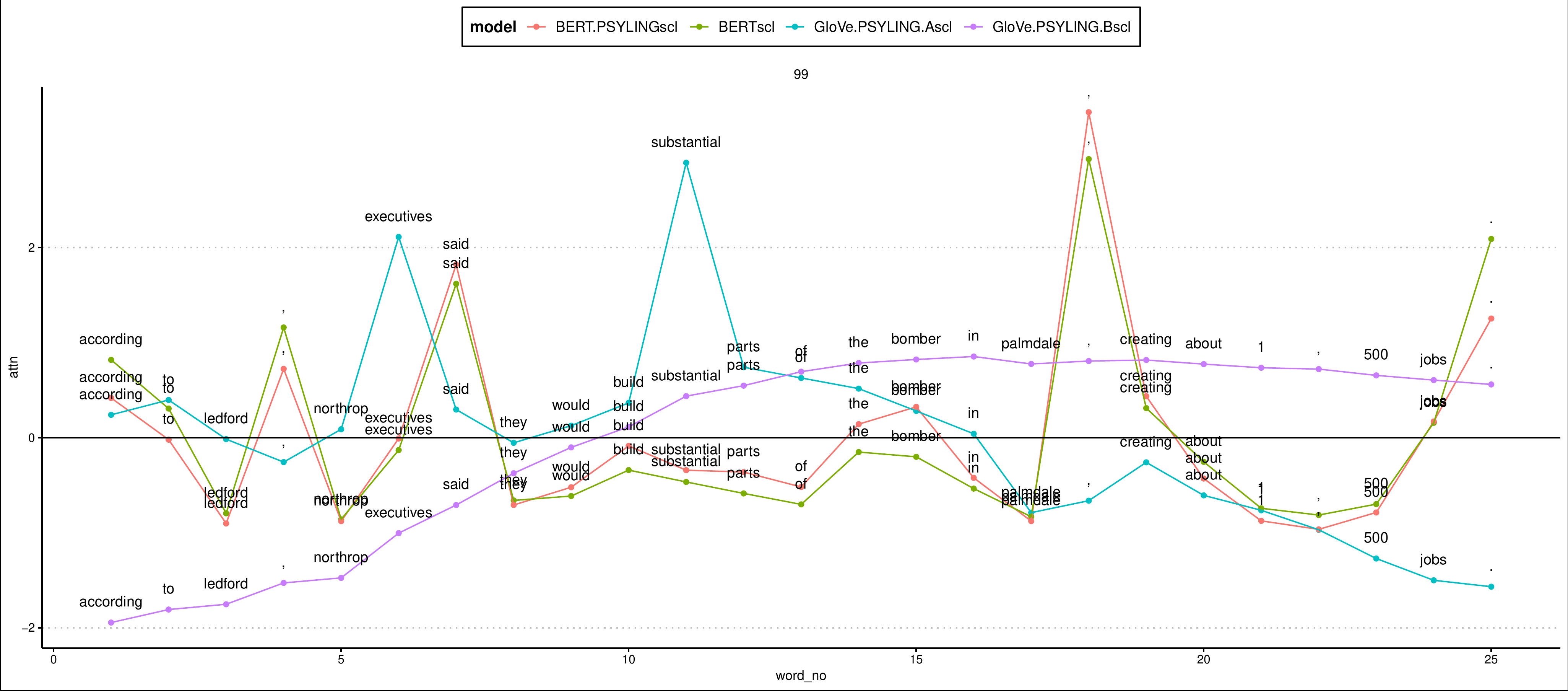}
    \caption{\textbf{Complexity explanation:} Distributions of attention weights over words in a randomly selected sentence.}
    \label{fig:attention}
\end{figure*}

\begin{table*}[]
\caption{\textbf{Simplification Generation:} Example pair from WikiLarge corpus (normal, simplified) and source sentence simplified by ACCESS model (including four parameter tokens) and ACCESS-XL (including ten parameter tokens).}
\label{tab:example}
\begin{tabularx}{\textwidth}{lX}
\hline
Type      & Sentence \\
\hline
Source (Wikipedia)   & One side of the armed conflicts is composed mainly of the   Sudanese military and the Janjaweed, a Sudanese militia group recruited   mostly from the Afro-Arab Abbala tribes of the northern Rizeigat region in  Sudan.\\
Target (WikiSimple)    & One side of the armed conflicts is made of Sudanese military   and the Janjaweed, a Sudanese militia recruited from the Afro-Arab Abbala tribes of the northern Rizeigat region in Sudan.\\
ACCESS    & One side of the armed conflict is made up of the Sudanese military and   the Janjaweed, a Sudanese militia group brought mostly from the Afro-Arab Abbala tribes of the northern Rizeigat region in Sudan.\\
ACCESS-XL & The army of the armed conflicts is mainly made of the Sudanese military and the Janjaweed, a Sudanese militia group. They recruited mostly from the   Afro-Arab Abbala tribes of the northern Rizeigat region in Sudan.\\
\hline
\end{tabularx}

\label{tab:example}
\end{table*}

\begin{figure*}
    \centering
    \includegraphics[width = 1\textwidth]{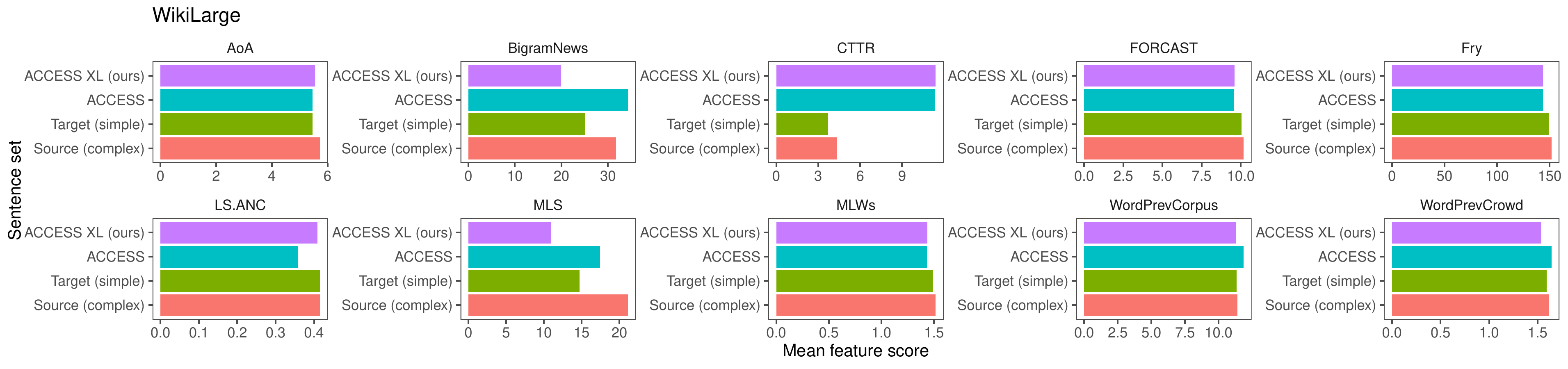}
    \includegraphics[width = 1\textwidth]{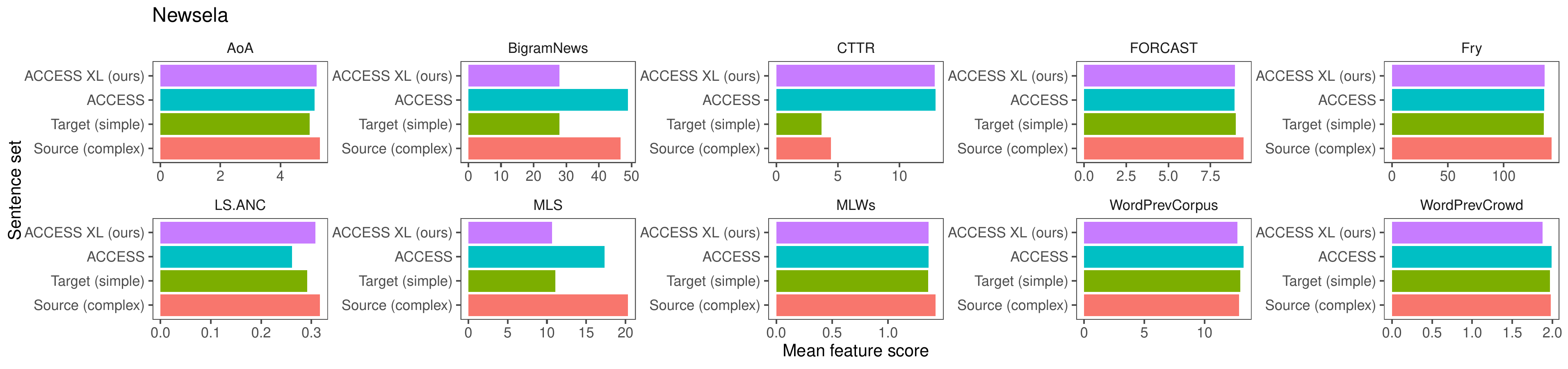}
    \includegraphics[width = 1\textwidth]{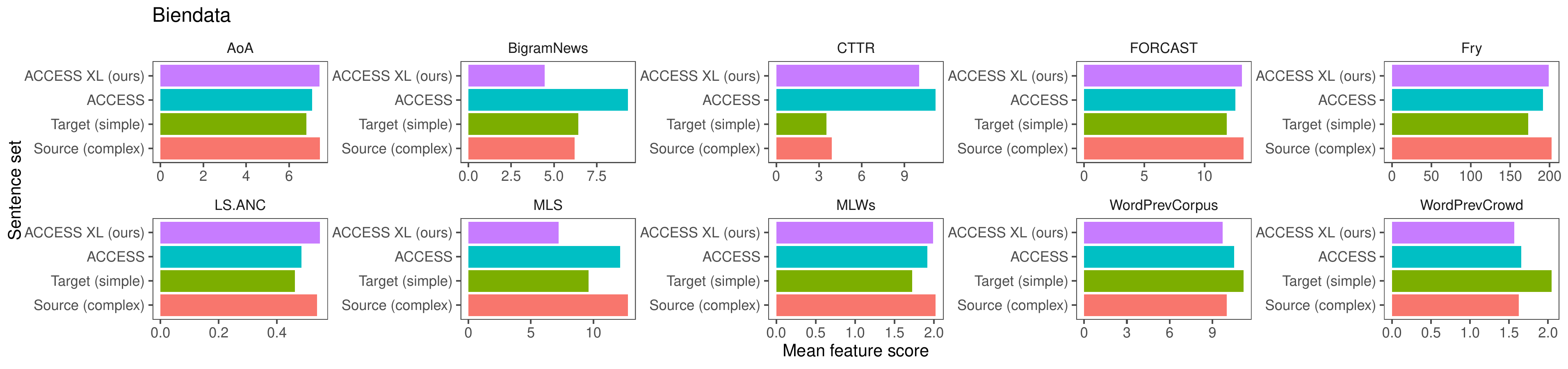}
    \caption{\textbf{Simplification Generation:} Mean values of the ten parameter tokens (engineered language features) across sentences sets.}
    \label{fig:behavior}
\end{figure*}

\begin{table*}[]
    \centering
        \caption{ACCESS model performance with prior complexity prediction using different complexity prediction models.}
    \label{tab:simp_results}
    \begin{tabular}{|l|l|r|r|r|r|}
        \hline
        ~ & ~ & \multicolumn{4}{c|}{ACCESS}\\ 
        ~ & ~ & \multicolumn{2}{c}{Ours} & \multicolumn{2}{c|}{Martin} \\ 
        \hline
        Dataset & Filter & SARI & FKGL & SARI & FKGL \\ 
        \hline
        \multirow{4}{*}{WiKiLarge} & BERT & 43.01 & 5.14 & 40.97 & 7.21 \\
        & BERT\_PSYLING & 42.84 & 5.06 & 40.97 & 7.17 \\
        & GloVe-PSYLING-a & 41.38 & 5.19 & 39.54 & 7.24 \\
        & GloVe-PSYLING-b & 41.53 & 5.03 & 39.72 & 7.22 \\
        \hline
        \multirow{4}{*}{Biendata} & BERT & 26.92 & 10.85 & 19.93 & 12.61 \\
        & BERT\_PSYLING & 26.87 & 10.86 & 19.87 & 12.63 \\
        & GloVe-PSYLING-a & 26.16 & 11.17 & 19.31 & 12.78 \\
        & GloVe-PSYLING-b & 26.87 & 10.90 & 19.89 & 12.62 \\
        \hline
        \multirow{4}{*}{Newsela} & bert & 33.44 & 5.27 & 27.33 & 6.78 \\
        & BERT\_PSYLING & 33.13 & 5.19 & 27.30 & 6.75 \\
        & GloVe-PSYLING-a & 34.88 & 3.96 & 29.41 & 6.45 \\
        & GloVe-PSYLING-b & 34.90 & 3.96 & 29.43 & 6.45 \\
        \hline
    \end{tabular}

    \label{accessPred}
\end{table*}

\begin{table}
    \centering
       \caption{Average ED between simple sentences and original ACCESS output predictions with and without complexity prediction. ED are calculated using the tseval library, which EASSE relies on.}
    \begin{tabular}{|l|l|l|}
    \hline
        Dataset&Filter&ED\\ \hline
       \multirow{6}{*}{WiKiLarge} & none & 15.641 \\
        & BERT & 15.566 \\ 
        & BERT\_PSYLING & 15.590 \\
        & GloVe-PSYLING\_a & 15.717 \\
        & GloVe-PSYLING\_b & 15.771 \\
        & LSTM & 15.705 \\ \hline
        \multirow{6}{*}{biendata} & none & 13.298 \\
        & BERT & 13.269 \\ 
        & BERT\_PSYLING & 13.267 \\ 
        & GloVe-PSYLING\_a & 13.240 \\
        & GloVe-PSYLING\_b & 13.281 \\
        & LSTM & 13.220 \\ \hline
        \multirow{6}{*}{newsela} & none & 16.378 \\
        & BERT & 15.958 \\
        & BERT\_PSYLING & 15.957 \\
        & GloVe-PSYLING\_a & 16.377 \\
        & GloVe-PSYLING\_b & 16.376 \\
        & LSTM & 16.008 \\ \hline
    \end{tabular}
 
    \label{evalViaED}
\end{table}

\begin{table}
    \centering
        \caption{Avg ED between complex sentences and original ACCESS outputs with/without complexity prediction}
    \begin{tabular}{|l|l|l|}
    \hline
        Dataset&Filter&ED\\ \hline
       \multirow{6}{*}{WiKiLarge} & none & 6.684 \\
        & BERT & 5.916 \\ 
        & BERT\_PSYLING & 5.979 \\
        & GloVe-PSYLING\_a & 5.639 \\
        & GloVe-PSYLING\_b & 5.765 \\
        & LSTM & 4.516 \\ \hline
        \multirow{6}{*}{biendata} & none & 2.823 \\
        & bert & 2.719 \\ 
        & BERT\_PSYLING & 2.699 \\ 
        & GloVe-PSYLING\_a & 2.529 \\
        & GloVe-PSYLING\_b & 2.723 \\
        & LSTM & 1.585 \\ \hline
        \multirow{6}{*}{newsela} & none & 5.368 \\
        & BERT & 3.918 \\
        & BERT\_PSYLING & 3.960 \\
        & GloVe-PSYLING\_a & 5.358 \\
        & GloVe-PSYLING\_b & 5.363 \\
        & LSTM & 3.022 \\ \hline
    \end{tabular}

    \label{evalViaEDcomplex}
\end{table}

\begin{table}[]
\caption{\textbf{Simplification Generation:} Proportion of sentences retained after complexity prediction after complexity prediction (step 1) across prediction model and dataset}
\setlength{\tabcolsep}{1pt}
\begin{tabular}{lcccc}
\hline
          &            \multicolumn{4}{c}{Complexity prediction model} \\
Dataset  &BERT             & PsyBERT          & PsyGloVe\textsubscript{a}         & PsyGloVe\textsubscript{b}         \\
\hline
Biendata        & 0.947          & 0.945            & 0.881               & 0.961               \\
Newsela        & 0.572          & 0.578            & 0.959               & 0.991               \\
WiKiLarge      & 0.807          & 0.805            & 0.675               & 0.764\\
\hline
\end{tabular}

\label{propRetained}
\bigskip
{\raggedright \textbf{Evaluation metrics for simplification generation}

FKGL is computed as a linear combination of the number of words per simple sentence and the number of syllables per word:

\begin{equation*}
FKGL = 0.39 \frac{N\ word}{N\ sent} + 11.8 \frac{N\ syl}{N\ word} - 15.59
\end{equation*}

SARI compares the predicted simplification
with both the source and the target reference. It is an average of F1 scores for three n-gram operations: additions ($add$), keeps ($keep$) and deletions ($del$). For each operation, these scores are then averaged for all n-gram orders (from 1 to 4) to get the overall F1 score.

\begin{equation*}
f_{ope}(n) = \frac{2 \times p_{ope}(n) \times r_{ope}(n)}{p_{ope}(n) + r_{ope}(n)}
\end{equation*}

\begin{equation*}
F_{ope} =\frac{1}{k} \sum\limits_{n = [1, ..., k]} f_{ope}(n)
\end{equation*}

\begin{equation*}
SARI =\frac{F_{add}+F_{keep}+F_{del}}{3} 
\end{equation*}

SARI thus rewards models for adding n-grams that occur in the reference but not in the input, for keeping n-grams both in the output and in the reference, and for not over-deleting n-grams. \citet{xu2016optimizing} show that SARI correlates with human judgments of simplicity gain.}
\end{table}

\begin{figure*}
    \centering
    \includegraphics[width= 1\textwidth]{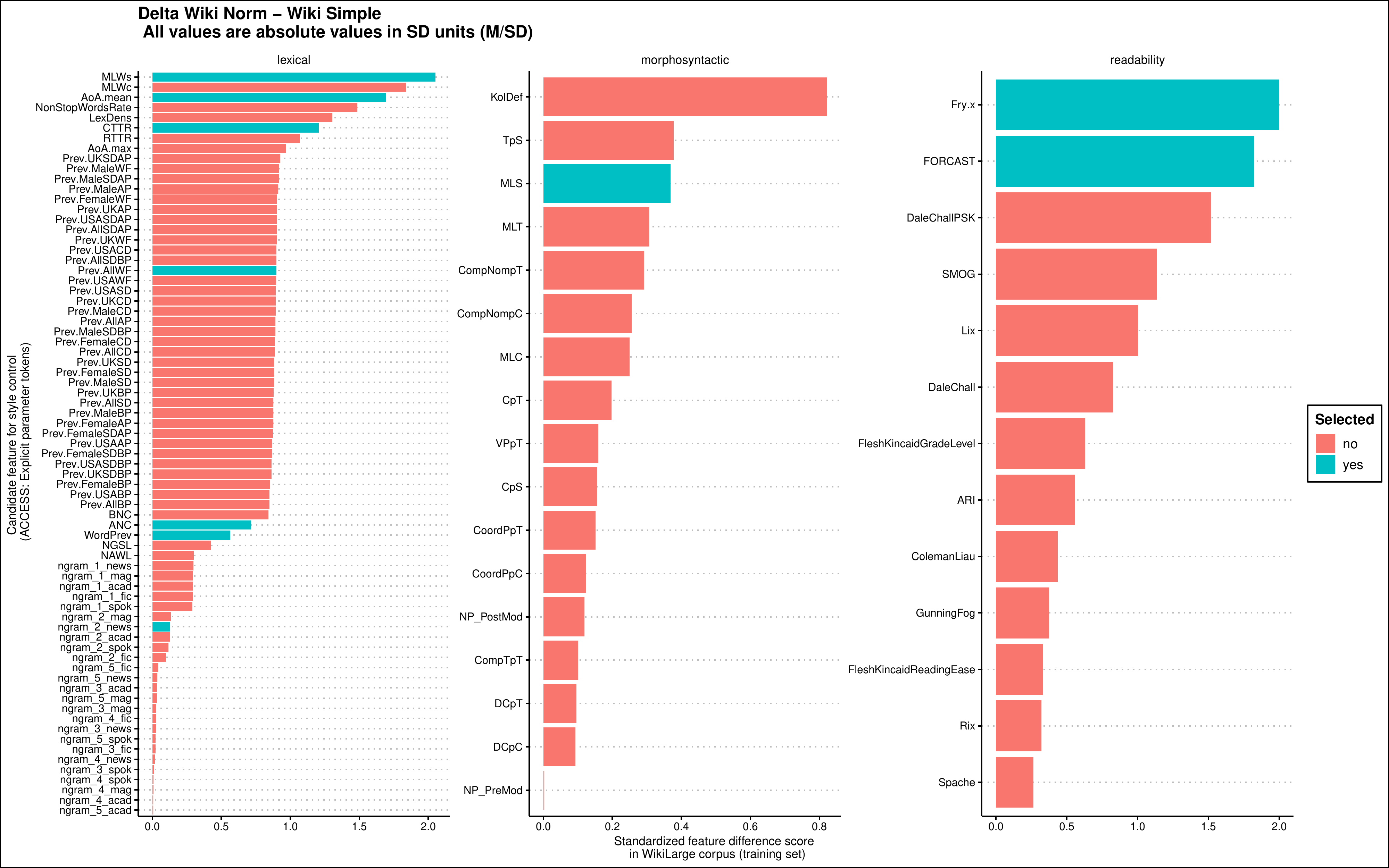}
    \caption{\textbf{Simplification Generation: }Differences in mean feature scores (standardized) between `normal' and `simple' sentences in WikiLarge corpus. Features in blue were selected for controllable sentence simplification in the ACCESS-XL model.}
    \label{fig:featureSelection}
\end{figure*}

\begin{figure*}
    \centering
    \includegraphics[width = 1\textwidth]{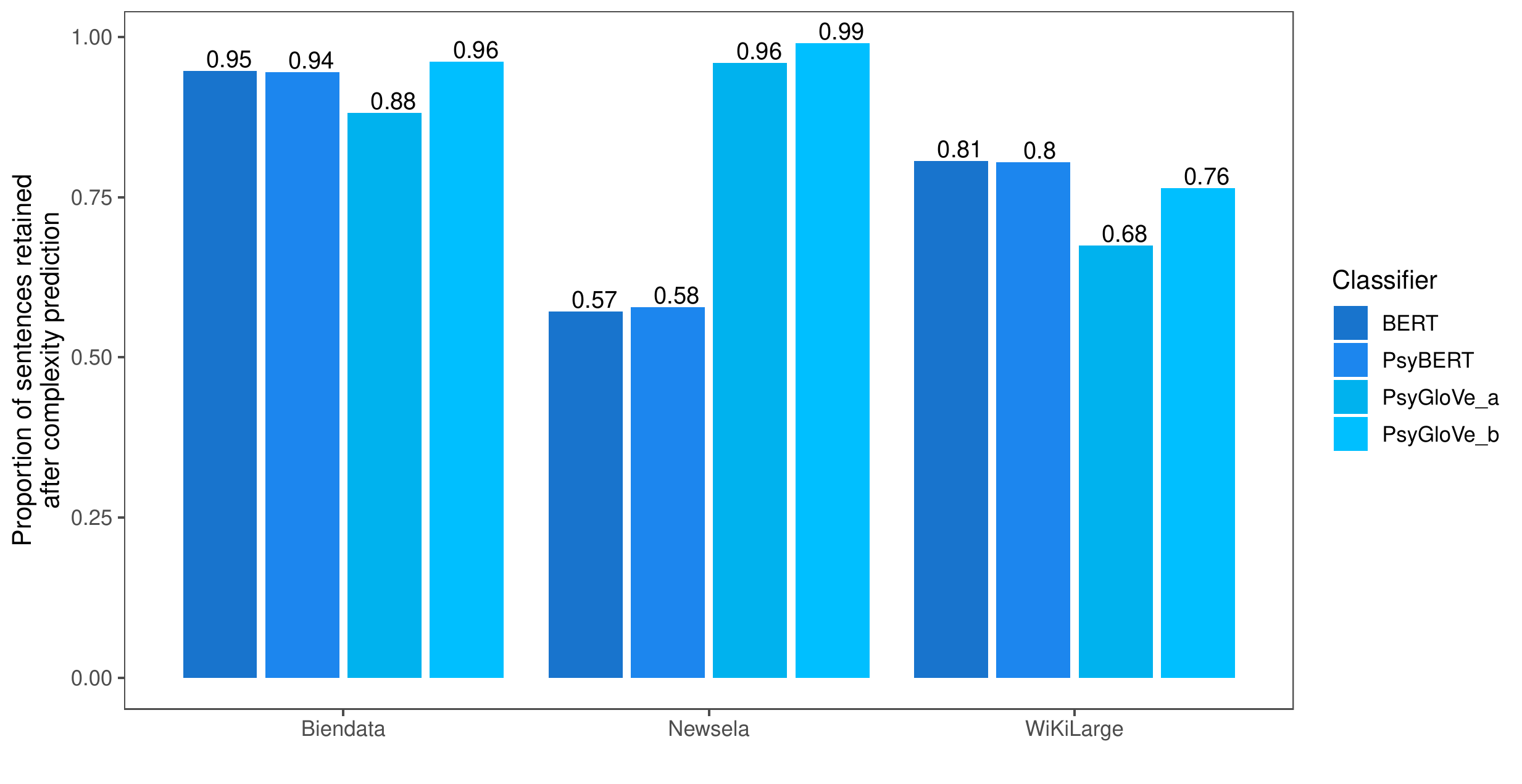}
    \caption{\textbf{Simplification Generation:} Proportion of sentences retained after complexity prediction after complexity prediction (step 1) across prediction model and dataset}
    \label{fig:propRetaineFig}
\end{figure*}

\begin{figure*}
    \centering
    \includegraphics[width = 1\textwidth]{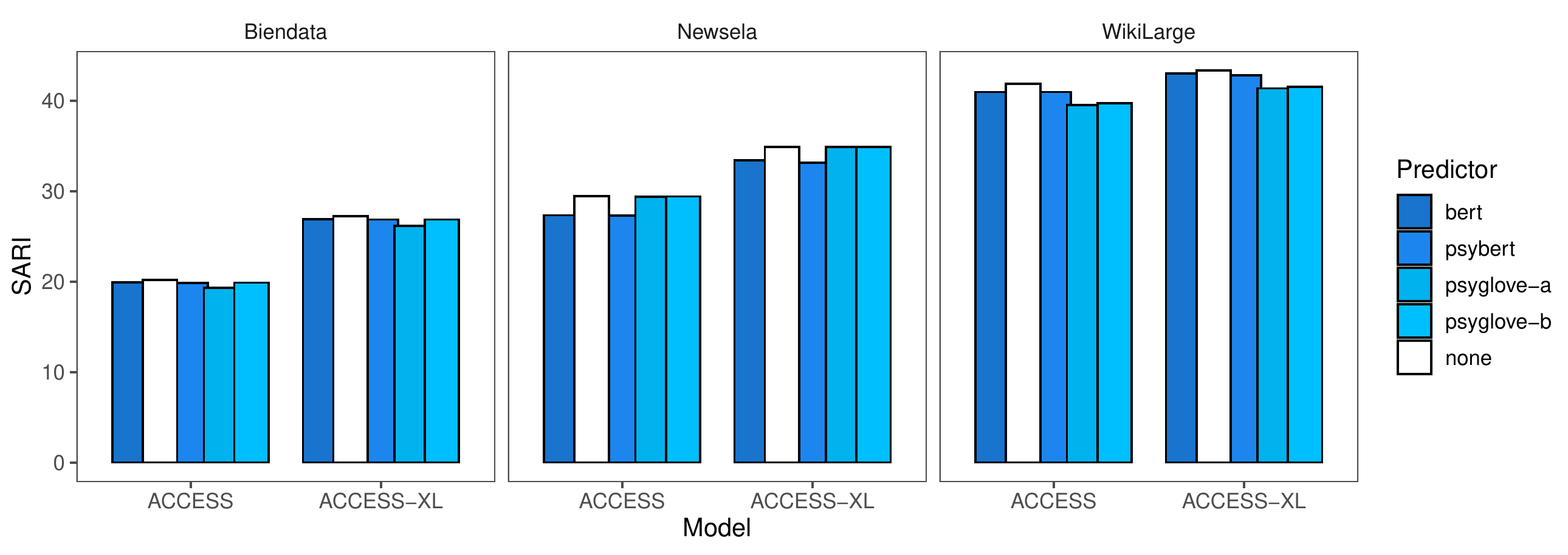}
    \includegraphics[width = 1\textwidth]{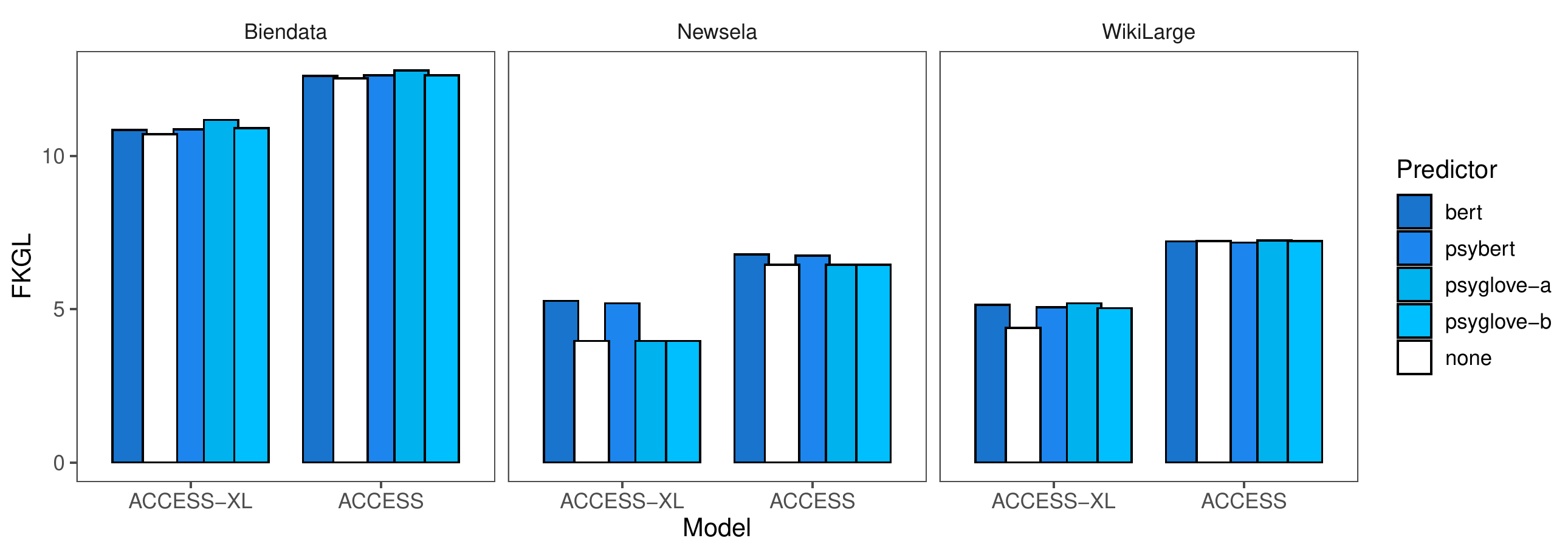}
    \caption{\textbf{Simplification Generation:} Performance of text simplification models as measured by SARI (top, higher is better) and Flesch-Kinkaid Grade Level (FKGL, bottom; lower is better) across datasets and use of complexity prediction methods.}
    \label{fig:sari}
    
\end{figure*}

\end{document}